\documentclass[runningheads]{llncs}
\usepackage[paperheight=235mm,paperwidth=155mm,textwidth=12.2cm,textheight=19.3cm]{geometry}
\usepackage{bbding}
\usepackage{graphicx}
\usepackage{listings}
\usepackage[T1]{fontenc}
\usepackage{pgfplots}
\pgfplotsset{width=10cm,compat=1.9}
\usepackage{tikz}
\usetikzlibrary{matrix,shapes,arrows,positioning}
\usepackage{caption}
\usepackage[backend=biber,style=numeric]{biblatex}
\makeatletter
\RequirePackage[bookmarks,unicode,colorlinks=true]  {hyperref}%
   \def\@citecolor{blue}%
   \def\@urlcolor{blue}%
   \def\@linkcolor{blue}%

\def\orcidID#1{\smash{\href{http://orcid.org/#1}{\protect\raisebox{-1.25pt}{\protect\includegraphics{orcid_color.eps}}}}}
\makeatother

\begin{document}

\title{LF-checker: Machine Learning Acceleration of Bounded Model Checking for Concurrency Verification (Competition Contribution)}

\titlerunning{Machine Learning Acceleration of Bounded Model Checking}
%
\author{Tong Wu\inst{1}\and
Edoardo Manino\inst{1}\and 
Fatimah Aljaafari\inst{1}\and 
Pavlos Petoumenos\inst{1}\and 
Lucas C. Cordeiro\inst{1}(\Envelope)}

\authorrunning{Tong Wu et al.}
%
\institute{University of Manchester, Manchester, United Kingdom
\email{lucas.cordeiro@manchester.ac.uk}
\\
}
\maketitle              
\begin{abstract}

We describe and evaluate LF-checker, a metaverifier tool based on machine learning. It extracts multiple features of the program under test and predicts the optimal configuration (flags) of a bounded model checker with a decision tree. Our current work is specialised in concurrency verification and employs ESBMC as a back-end verification engine. In the paper, we demonstrate that LF-checker achieves better results than the default configuration of the underlying verification engine.

\end{abstract}

\section{Overview}

Bounded Model Checking (BMC) has been successfully applied to prove whether a piece of software meets the expected requirements. However, when the software features concurrent execution, the main challenge becomes exploring the exponentially growing program state space. For this reason, there have been many efforts to improve BMC for concurrent software, including sequentialization~\cite{DBLP:conf/cav/InversoT0TP14}, dedicated theories for multi-threaded programs~\cite{DBLP:conf/pldi/HeSF21} or combining BMC with fuzzing~\cite{DBLP:journals/access/AljaafariMMSMC22}. A state-of-the-art example of such efforts is ESBMC, an efficient SMT-based bounded model checker for C and C++ programs~\cite{esbmc}. More specifically, ESBMC explores each thread interleaving up to a maximum number of context switches~\cite{DBLP:conf/icse/CordeiroF11}.

In order to control the program states explosion, BMC tools like ESBMC can be executed with different settings (flags). These flags control the various mathematical techniques used by the BMC tool. Unfortunately, expert knowledge is required to set the optimal flags for a given concurrent program. As a consequence, BMC tools are often executed with their default settings, which leads to compromises in their performance. 

In this paper, we introduce the metaverifier tool LF-checker (learn-from-the-checker), which is able to predict the optimal flags for a given concurrent program. Specifically, LF-checker uses ESBMC~\cite{esbmc} as a back-end verifier, and predicts its optimal settings with a decision tree. The decision tree is trained to recognise relevant features of the input C program. Overall, LF-checker outputs $34\%$ more correct results than the default setting of ESBMC in the \texttt{Concurrency-} \texttt{Safety} category of SV-COMP 2023.

\section{Software Architecture}

\begin{figure}[t]
\begin{tikzpicture}[node distance=2cm]
\tikzstyle{io} = [rectangle, rounded corners, minimum width=1cm, minimum height=1cm, text centered, draw=black, fill=black!20]
\tikzstyle{obj} = [rectangle, rounded corners, minimum width=1.5cm, minimum height=1cm, text centered, draw=black]
\tikzstyle{arrow} = [thick,->,>=stealth]
\node (ccode) [io]{\scriptsize C Code};
\node (frontend) [obj, right of = ccode, xshift=0.5cm] {\scriptsize feature extractor}; 
\node (flaggenerator) [obj, right of =frontend,  xshift=0.5cm]{\scriptsize flag generator};
\node (model) [obj, below of= flaggenerator]{\scriptsize ML Model}; 
\node (esbmc) [obj, right of= flaggenerator, xshift = 1.5cm]{\scriptsize ESBMC}; 
\node (property) [io, right of=esbmc, xshift=0.3cm]{\scriptsize Property File};
\node (out) [io, below of=property]{\scriptsize Result};
\draw[->] (ccode.east) -- (frontend.west);
\draw[->] (frontend.south) -- node[anchor=east]{\scriptsize extract features}(model.west);
\draw[->] (model.north) -- node[anchor=west]{\scriptsize best prediction} (flaggenerator.south);
\draw[->] (flaggenerator.east) -- node[anchor=south]{\scriptsize optimal flags} (esbmc.west);
\draw[->] (property.west) -- (esbmc.east);
\draw[->] (esbmc.south) -- (out.west);
\end{tikzpicture}
\caption{Workflow of LF-checker. The grey boxes represent the inputs and outputs, and the white
boxes represent the running process.} 

\label{fig2}
\end{figure}
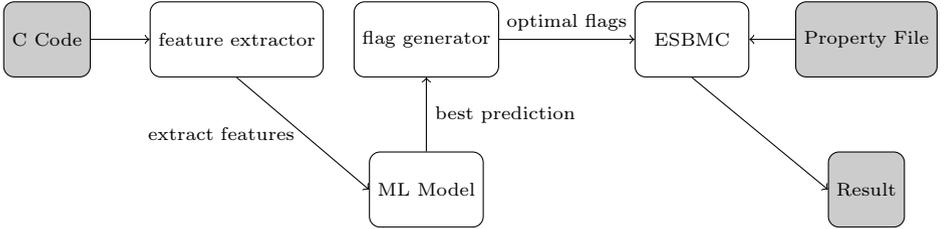

The general workflow of LF-checker is shown in Fig.~\ref{fig2}. First, we analyse the source code of the program-under-test (PUT), and extract useful features. Then, we pass these features to the ML model, which predicts the best flags for ESBMC. Finally, we run ESBMC on the PUT with these flags, with the objective of proving or disproving a given safety property. Since our contribution is centered around the ML model, we share the details of its design below.

\paragraph{Back-end Verification Engine (ESBMC).} ESBMC exposes a large number of flags that regulate its verification strategy for concurrent program. We list them in Table~\ref{Tab3} together with their type and default values. 

\paragraph{Feature Engineering.} We compute the abstract syntax tree of the PUT~\cite{DBLP:journals/sigsoft/NeamtiuFH05}, and extract all possible verification-relevant features from it. For the sake of brevity, we list them in Table~\ref{Tab2}.

\paragraph{Training Set Labeling.} The training set contains 20\% of the benchmarks in the \texttt{unreach-call.ConcurrencySafety} category in SV-COMP 2023. \textcolor{black}{For each of the 153 training benchmarks,} we run ESBMC for 3 minutes with \textcolor{black}{240} different combinations of flags. \textcolor{black}{These flag combinations were chosen among those that improved the verification result the most during our preliminary experiments.} We summarise the verification time and verdict for each of the $36720$ training samples with an ordinal label ranging from $0$ to $5$, according to the classification in Table~\ref{Tab1}. Lower values of the output class are more desirable, as they correspond to fast and correct verdicts. 

\paragraph{Machine Learning Model.} We train a decision tree classifier~\cite{song2015decision} in a supervised fashion using \texttt{Scikit-learn}~\cite{scikit-learn}. \textcolor{black}{We balance the training set by weighing each sample by its class frequency. Furthermore, we set the splitting threshold for internal nodes to 4 samples, and the minimum number of samples in a leaf nodes to 3.} The decision tree learns to predict the $0\!-\!5$ output class given the features of the PUT and a given choice of flags. 

\paragraph{Runtime Flag Prediction.} At runtime, we use the decision tree to predict the optimal set of flags for ESBMC. \textcolor{black}{More specifically, we try all $240$ flag combinations and pick the one that yields the lowest output class. Since decision trees are very fast, this process takes only a few seconds to run.}

\begin{table}[t]
\begin{minipage}{.34\linewidth}
\begin{center}  
\resizebox{\textwidth}{!}{
\begin{tabular}{||l|l||}
\hline
 \textit{Verification Result} & \textit{Class} \\
 \hline
 correct \& time <= 10 (s)  & 0 \\  
 correct \& 10 (s) < time < 60 (s)  & 1 \\ 
 correct \& time >= 60 (s)  & 2 \\ 
 unknown & 3\\
 timeout & 4\\
 incorrect & 5\\
 \hline
\end{tabular}}
\captionsetup{skip=5pt}
\caption{\scriptsize Labeling result classes}
\scriptsize \label{Tab1}
\end{center}
\end{minipage}\hfill
\begin{minipage}{.29\linewidth}
\begin{center}
\resizebox{\textwidth}{!}{
\begin{tabular}{||l||}
\hline
 \textit{Program Features}\\
 \hline
 threads created\\  
 threads joined \\ 
 mutex locks \\ 
 atomic locks \\
 global variables access \\
 global function called \\
 binary operators \\
 nondet variables \\
 min. global variable access times  \\
 min. global function called times   \\
 iteration times in loop  \\
 \hline
\end{tabular}}
\captionsetup{skip=5pt}
\caption{\scriptsize Program features 
}
\label{Tab2}
\end{center}
\end{minipage}\hfill
\begin{minipage}{.36\linewidth}
\begin{center}
\resizebox{\textwidth}{!}{
\begin{tabular}{||l |l | l||}
\hline
\textit{ESBMC Flags} & \textit{Type} & \textit{Default}\\
\hline
     context-bound &  Integer & unlimited\\
     strategy& String & None\\
     k-step & Integer & 1\\
     unwind & Integer & unlimited\\
     no-por & Boolean & disabled\\
     no-goto-merge & Boolean & disabled\\
     state-hashing & Boolean & disabled\\
     add-symex-value-sets & Boolean & disabled\\
\hline
\end{tabular}}
\captionsetup{skip=5pt}
\caption{\scriptsize ESBMC flags}
\label{Tab3}
\end{center}
\end{minipage}
\end{table}

\section{Strengths and Weaknesses}

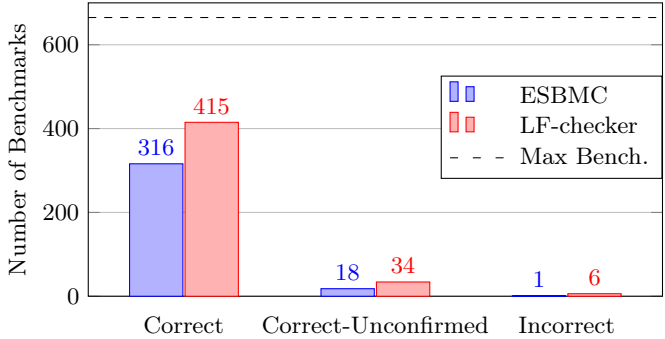
\begin{figure}[t]
\centering
\begin{tikzpicture}
\begin{axis}[
        width  = 0.75*\textwidth,
        height = 0.45*\textwidth,
        major x tick style = transparent,
        ybar=2*\pgflinewidth,
        bar width=20pt,
        ymajorgrids = true,
        ylabel = {Number of Benchmarks},
        symbolic x coords={Correct,Correct-Unconfirmed,Incorrect},
        xtick = data,
        scaled y ticks = false,
        enlarge x limits=0.25,
        ymin=0,
        ymax=700,
        legend cell align=left,
        legend style={
                at={(1,0.4)},
                anchor=south east,
                column sep=1ex
        },
        nodes near coords,
        point meta=y
    ]
    \addplot coordinates {(Incorrect,1) (Correct-Unconfirmed,18) (Correct,316)};
    \addplot coordinates {(Incorrect,6) (Correct-Unconfirmed,34) (Correct,415)};
    \legend{ESBMC, LF-checker}
    
    \coordinate (A) at (axis cs:Correct,665);
    \coordinate (O1) at (rel axis cs:0,0);
    \coordinate (O2) at (rel axis cs:1,0);
    \draw [black,sharp plot,dashed] (A -| O1) -- (A -| O2);
    \addlegendimage{line legend,dashed,black}
    \addlegendentry{Max Bench.}
    
\end{axis}
\end{tikzpicture}
\caption{Comparison between the verification results of ESBMC (default settings) and LF-checker. The tools' verdicts are grouped according to their correctness.}
\label{fig1}
\end{figure}

\textcolor{black}{Figure~\ref{fig1} summarises the verification results of the} \texttt{unreach-call.Concurrency-} \texttt{Safety} category \textcolor{black}{of SV-COMP 2023. More in detail,} ESBMC outputs 334 correct results including 18 unconfirmed, and 1 incorrect verdict. In contrast, LF-checker outputs 449 correct results including 34 unconfirmed, as well as 6 incorrect verdicts. This shows that, \textcolor{black}{by optimising the flags of ESBMC for different programs with our predictor,} ESBMC can output 34\% more correct results compared to using its default settings.

However, LF-checker also outputs 5 additional wrong results. This is because of an issue in the latest version of ESBMC which leads to wrong verdicts in a small number of concurrent benchmarks \textcolor{black}{when we merge goto statements after a context switch}. 
Crucially, these benchmarks have similar program features to those that are handled correctly by ESBMC, which leaves no opportunity for the ML model to filter them out. 

Besides, \textcolor{black}{when many flag combinations are predicted in the same output class,} we naively choose the first one. \textcolor{black}{A more granular predictor may be able to discriminate between them, and achieve better results. We leave this research question to future work.} Finally, our model is only trained on the \texttt{unreach-call.Concurren-} \texttt{cySafety} category. For the other subcategories of \texttt{ConcurrencySafety}, which were newly introduced in SV-COMP 2023, we apply the default settings of ESBMC. As our experiments show, this choice is far from optimal.

\section{Tool Setup and Configuration}
The competition submission is based on LF-checker version 1.0 \footnote{\url{https://github.com/Anthonysdu/lf-checker}} and the back-end ESBMC version 7.0.0 64-bit x86\_64 linux \footnote{\url{https://github.com/esbmc/esbmc}}. LF-checker is called by executing the script \textsf{esbmc-wrapper.py}. It reads a \textsf{.c} or \textsf{.i} file and the property file as well as trained ML model as inputs. As an example, we can run the tool by executing the following command:
\\\\
\textsf{./esbmc-wrapper.py -p propertyFile --concurrencyFlagPredictor model.sav --arch 32 benchmark}\\\\
\sloppy where \textsf{model.sav} indicates the trained ML model file. We reuse the same Benchexec tool info module as ESBMC which is called \textsf{esbmc.py} and the benchmark definition file is \textsf{lf-checker.xml}. The python packages \textsf{anytree} and \textsf{scikit-learn} should be installed separately in the SV-COMP machines.
\section{Software Project and Contributors}

Tong Wu maintains LF-checker. It is publicly available under a BSD-style license. The source code is available at \url{https://github.com/Anthonysdu/lf-checker}, and instructions for running the tool are given in the \textsf{README} file.

\section*{Data-Availability Statement}
The project archive for SV-COMP 2023 is available at Zenodo~\cite{wu_2022}.
\section*{Acknowledgment}
The work in this paper is partially funded by the EPSRC grants EP/T026995/1, EP/V000497/1, EU H2020 ELEGANT 957286, and Soteria project awarded by the UK Research and Innovation for the Digital Security by Design (DSbD) Programme. \textcolor{black}{We also would like to acknowledge Mengze Li and Huajie He, who inspired our work.}

\printbibliography
\end{document}